\documentclass[11pt,a4paper]{article}
\usepackage{times,latexsym}
\usepackage{url}
\usepackage[T1]{fontenc}
\usepackage{booktabs}
\usepackage{graphicx}
\usepackage{subfigure}
\usepackage{caption}
\usepackage{subcaption}
\usepackage{amsthm}
\usepackage{multirow}
\theoremstyle{definition}

\usepackage{listings}
\usepackage[table,xcdraw]{xcolor}
\usepackage[acceptedWithA]{tacl2021v1}

\usepackage{xspace,mfirstuc,tabulary}

\newif\iftaclinstructions
\taclinstructionsfalse 
\iftaclinstructions
\newcommand{\instr}
\fi

\iftaclpubformat 

\else

\fi

\title{PUB: A \underline{P}ragmatics \underline{U}nderstanding \underline{B}enchmark for Assessing LLMs' Pragmatics Capabilities} 

\author{
  Settaluri Lakshmi Sravanthi$^\diamond$\Thanks{Equal contribution in coding and experiments}, Meet Doshi$^\diamond$\footnotemark[1], Tankala Pavan Kalyan$^\diamond$\footnotemark[1],\
  \\
\textbf{Pushpak Bhattacharyya$^\diamond$, Rudra Murthy$^\S$, Raj Dabre$^\ddagger$}
  \\
  $^\diamond$CFILT, Indian Institute of Technology Bombay
  \\
  $^\S$IBM Research
  \\
  $^\ddagger$NICT, Japan
  \\
  \texttt{\{sravanthi,meetdoshi,pb\}@cse.iitb.ac.in}, 
  \\
  \texttt{190020124@iitb.ac.in, rmurthyv@in.ibm.com, prajdabre@gmail.com}
}

\date{}
\begin{document}
\maketitle
\begin{abstract}
LLMs have demonstrated remarkable capability for understanding semantics, but they often struggle with understanding pragmatics. To demonstrate this fact, we release a Pragmatics Understanding Benchmark (PUB) dataset consisting of \textit{fourteen} tasks in \textit{four} pragmatics phenomena, namely, {\it Implicature, Presupposition, Reference, and  Deixis}. We curated high-quality test sets for each task, consisting of Multiple Choice Question Answers (MCQA). PUB includes a total of $28k$ data points, $6.1k$ of which have been created by us, and the rest are adapted from existing datasets. We evaluated \textit{nine} models varying in the number of parameters and type of training. Our study indicates that fine-tuning for instruction-following and chat significantly enhances the pragmatics capabilities of smaller language models. However, for larger models, the base versions perform comparably with their chat-adapted counterparts. Additionally, there is a noticeable performance gap between human capabilities and model capabilities. Furthermore, unlike the consistent performance of humans across various tasks, the models demonstrate variability in their proficiency, with performance levels fluctuating due to different hints and the complexities of tasks within the same dataset. Overall, the benchmark aims to provide a comprehensive evaluation of LLM's ability to handle real-world language tasks that require pragmatic reasoning. 
\end{abstract}

\begin{figure}[h!]
    \centering
    \includegraphics[width=1\linewidth]{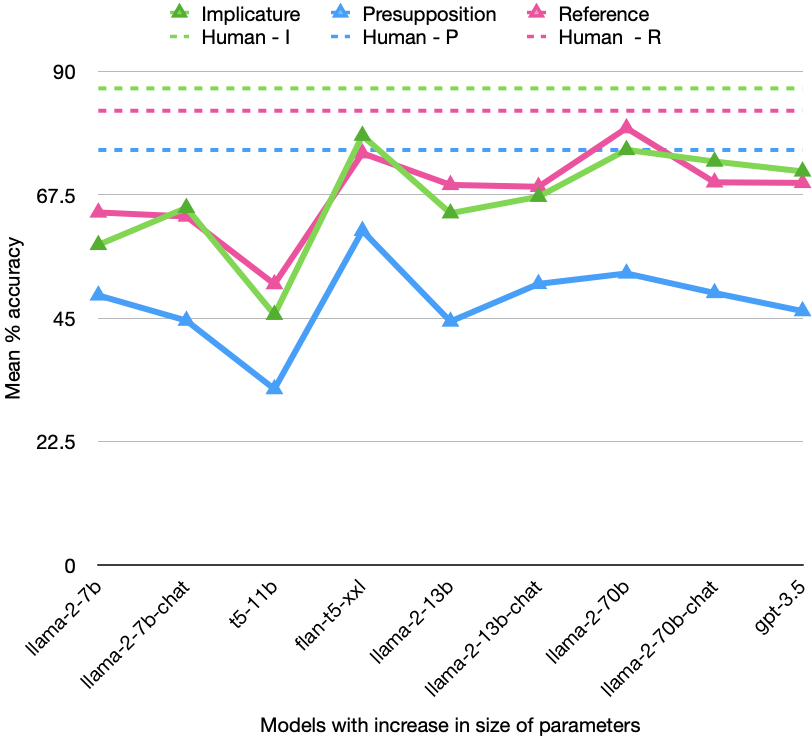}
    \caption{Average performance of models on three different pragmatics phenomena. Average accuracy for reference and deixis are merged and plotted as \textit{Reference} as they are closely related phenomena. Human - I, P, R represent the performance of human evaluators on Implicature, Presupposition, and Reference respectively}
    \label{performance}
\end{figure}
\section{Introduction}
Pragmatics, within linguistics, examines how context shapes language understanding in communication \citep{grice1975logic}. It centers on real-life language use, considering context, speaker intentions, presuppositions, and implied meanings to derive interpretations beyond literal words. Human's proficiency in pragmatics stems from their inherent cognitive skills and social awareness. Our minds adeptly process not only spoken words but also context and implied messages.

In the realm of Natural Language Processing (NLP), Large Language Models (LLMs) (GPT-3 \citep{DBLP:conf/nips/BrownMRSKDNSSAA20}, BLOOM \citep{bloom}, PaLM  \citep{palm},  LLAMA-2 \cite{llama-2}, others) have emerged as a transformative force in recent years. LLMs have shown remarkable abilities on many downstream tasks like NLU (GLUE \citep{glue},  MultiNLI \citep{multinli}), Text generation (LAMBADA, Wikitext), Code synthesis (APSS, HumanEval \citep{humaneval}), QA (Natural Questions, ARC, OpenbookQA \citep{openbookqa}, SQuAD \citep{squad}), Reasoning (SuperGLUE \cite{superglue}, GSM8k \citep{gsm8k}, Strategy QA \citep{strategyqa}), etc.

As LLM's capabilities have expanded, they are now being utilized in practical real-world applications like chatbots, search engines, and web browsers. Given the increased interaction between humans and LLMs, the following research questions need to be answered: \\
\noindent a. \textit{How much do LLMs understand what humans mean during conversations?}\\
\noindent b. \textit{Is there a correlation between a model's pragmatics abilities and its scale?}\\
\noindent c. \textit{Do LLMs that are optimized for dialogue use-cases exhibit superior pragmatic abilities?}\\
\noindent d. \textit{Despite operating on the same dataset, do LLMs demonstrate varying task sensitivity?}\\
\noindent e. \textit{How do the pragmatic abilities of LLMs compare concerning world knowledge involvement?}\\
\noindent f. \textit{Do they understand the same implied meaning and make the same assumptions as us?}

 To answer these questions we lean towards the domain of \textit{pragmatics}. While semantics involves the study of words and their meanings in a language, \textit{pragmatics} extends this inquiry by considering word's meanings within the context in which they are used. 
 Most benchmarks until now deal only with abilities like problem-solving \cite{gsm8k} or semantic understanding (GLUE \citep{glue}, BigBench \cite{bigbench}, etc.) where LLMs have started to come close or be at par with human benchmarks. Despite the recent progress, we notice that there is still a lot of pragmatic understanding gap between what the language model understands and what was meant by a statement. To facilitate this research, we propose a Pragmatic Understanding Benchmark (PUB) over four major Pragmatic phenomena, namely, Implicature (Understanding what is suggested or implied in a statement even though it is not literally expressed), Presupposition (An implicit assumption that is taken for granted before the use of a statement), Deixis (a phenomenon in which certain words or phrases within a sentence or discourse rely on contextual cues, such as the speaker, the listener, or the surrounding context, to convey their meaning effectively) and Reference (how language points to things, people, place, time, etc) in accordance with the content and structure outlined in the Handbook of Pragmatics  \cite{handbook}.

In PUB, we've constructed tasks based on datasets focusing on Implicature, Presupposition, Deixis and Reference. The benchmark includes 22,000 examples, leveraging existing data, and introduces three new datasets with 6,100 newly annotated examples. Human evaluation of a subset of these datasets is conducted to assess performance against established LLMs.  The benchmark comprises fourteen tasks that evaluate pragmatics as an MCQA task since MCQA evaluation is more closely related to question-answering abilities in conversations \cite{RobinsonW23}. We carefully curate the existing datasets to balance them and formulate prompts for these tasks, which are more natural and better suited to evaluate LLMs.
Following (\citep{DBLP:conf/nips/BrownMRSKDNSSAA20},  \citep{RobinsonW23}), we evaluate the pragmatic abilities of LLMs using Multiple Choice Prompting (MCP) and Cloze prompting (CP). To validate the model’s confidence in its choices we also calculate the Proportion of Plurality Agreement (PPA) 3 tasks similar to \citep{RobinsonW23}, this way we can evaluate the model’s certainty in its predictions to achieve higher performance. 

Our contributions are: (1) a comprehensive and unified dataset for 14 distinct tasks in pragmatics (Figure: \ref{dataset-diagram}), containing 28k data points; to the best of our knowledge this is the first dataset- linguistically motivated and well-grounded- to test pragmatic capabilities of LLMs\footnote{The benchmark is available at \url{https://huggingface.co/datasets/cfilt/PUB}}. (2) a systematic evaluation of 6 variations of llama-2, t5, Flan-t5, and GPT-3.5, on the 14 mentioned tasks. (3) a study of human performance on a sample of the dataset to highlight the performance gap between LLMs and humans. (4) insight emerging from (3) to uncover strengths and weaknesses of LLMs vis-a-vis humans. 
These contribution points- we hope- will assist researchers in improving the interactive abilities of LLMs.

\section{Related work}
Pragmatics is very crucial in the domain of linguistics, where it plays a critical role in understanding meaning \citep{moris}. In linguistic terms, pragmatics deals with the study of context-dependent aspects of meaning that are systematically abstracted away from, in the construction of content or logical form \cite{handbook}.  Some of the basic subfields of pragmatics include implicature, presupposition, speech acts, reference, deixis, definiteness, and indefiniteness. 
Over the years, many researchers have devoted their research to studying such pragmatic phenomena for machine learning. To study implicatures, \citet{circa} employ indirect answers in polar questions, \citet{grice} utilize hierarchical grammar models for understanding implicature and deictic reference in simple conversations, \citet{impres} employ Natural Language Inference (NLI) to grasp scalar implicatures, \citet{Implicature} leverage implicature rules for optimizing sentiment detection, and \citet{squinky} develop a sentence-level corpus with implicature ratings. Whereas for presupposition, \citet{qa2} use search engine queries that may contain questionable assumptions that are closely related to presupposition. \citet{Kabbara} also reveals that Transformer models exploit specific structural and lexical cues as opposed to performing some kind of pragmatic reasoning. 

Recent studies \citep{hp, ruis2023goldilocks} highlight language models' struggle with humor, irony, and conversational maxims. Previous evaluations either focused on singular phenomena or had limited sample sizes, like \citep{Implicature, pragmeval, pragmaticqa}. To the best of our knowledge, we are the first ones to combine major aspects of pragmatics to create a quantifiable benchmark.
\section{Datasets}
With the help of language experts, we selected existing datasets covering important pragmatic aspects. Specifically, we select Circa \citep{circa}, GRICE \citep{grice}, FigQA \citep{figqa}, FLUTE \citep{flute}, IMPPRES \citep{impres}, and NOPE \citep{nope}. We adapted datasets for various tasks (in MCQA format) with necessary changes and also made new ones where needed for specific purposes. Details of newly annotated datasets are discussed below:
\begin{enumerate}
    \item \label{CircaPlus} \textbf{CircaPlus} is our newly annotated dataset containing 2.5k human written implied meanings based on the indirect responses present in Circa dataset \citep{circa}. 
    \item \label{DialogAssumptions} \textbf{DialogAssumptions} is a new dataset containing 2.5k pairs of expert-annotated presuppositions based on a subset of dialogues from the Dailydialog dataset \citep{dailydialog}. While current presupposition datasets are built around trigger words present in sentences, to our understanding, there hasn't been a resource addressing presuppositions in conversational contexts where trigger words are absent. Hence, we developed this dataset specifically to fill this gap. 
    \item  \textbf{MetoQA} is a novel dataset comprising 1100 multiple-choice questions based on the linguistic phenomenon called metonymy. Metonymy is a figure of speech in which one word or phrase is substituted with another word or phrase with which it is closely associated or related. Unlike a metaphor, where one thing is said to be another (e.g., ``Life is a journey"), in metonymy, the substitution is based on a real, often contiguously related, connection between the two terms (e.g., ``These are my hired guns").
\end{enumerate}

\section{Tasks}
\begin{figure*}[h!]
    \centering
    \includegraphics[width=0.9\textwidth]{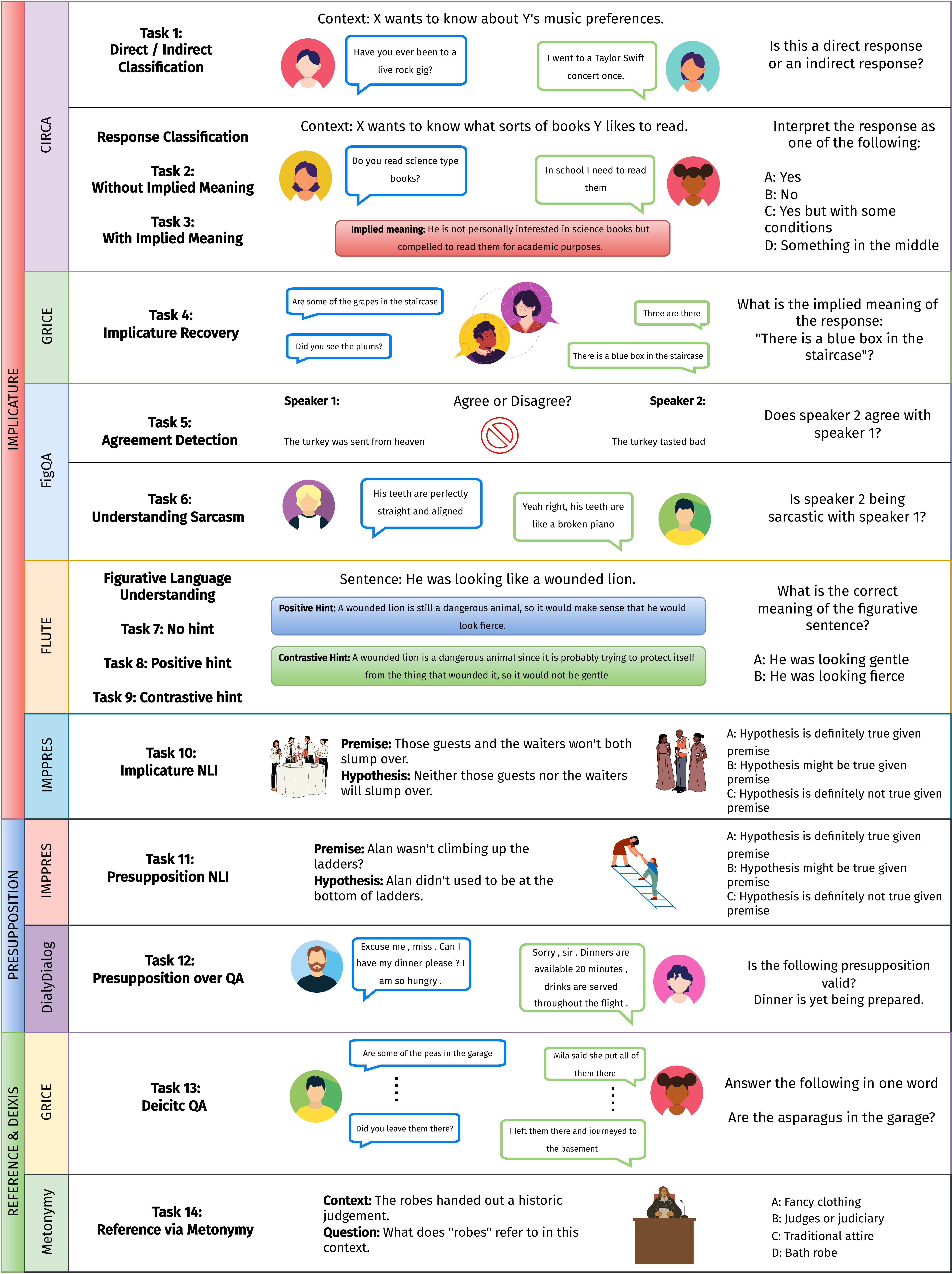}
    \caption{Examples of each task from PUB, The tasks are divided across \textit{four} domains of pragmatics (Implicature, Presupposition, Reference, and Deixis). Our proposed benchmark builds upon existing pragmatic datasets and combines our newly annotated datasets comprising 6k annotations to complete the pragmatic evaluation test suite with 28k examples. We have reformatted the existing datasets into MCQA prompts that explicitly test these abilities.}
    \label{dataset-diagram}
\end{figure*}
In this section, we describe each task and the associated dataset. Each task incorporated within PUB is structured to evaluate distinct domains of pragmatics.  Figure \ref{dataset-diagram} contains examples from each task. 
\subsection{Implicature}
Implicature, an unspoken aspect of a speaker's meaning, extends beyond the literal content in a speaker's message. Understanding implicature is crucial for LLMs, as it allows them to interpret context, discern implied messages, and produce responses that surpass literal text, ensuring more contextually suitable, human-like, and meaningful interactions. Owing to the importance of implicature in pragmatics we have designed \textit{ten} tasks that thoroughly test the LLM's abilities to capture this phenomenon.

\noindent{\textbf{Task 1 - Direct/Indirect classification}}
This task evaluates language models' capability to distinguish between direct and indirect responses, crucial for understanding user intentions in dialogue systems. The model receives context, a question, and a response (that can be direct or indirect) and then selects between two options: A) Direct answer and B) Indirect answer. We utilized a label-balanced set of 2,500 data points sourced from the Circa dataset for this purpose.

\noindent{\textbf{Task 2 and 3 - Response classification without implied meaning and with implied meaning:}} 
Task 2 involves categorizing indirect answers using five labels. The model receives context, a question, and an indirect answer and must choose the most fitting label from options A) Yes, B) No, C) Yes, subject to conditions, D) In the middle, neither yes nor no, E) Other. This task evaluates LLMs' ability to comprehend indirect responses, specifically within polar Question and Answer scenarios, utilizing the Circa dataset. Task 3, an extension of Task 2, introduces implied meanings as additional cues to assist LLMs in interpreting indirect answers. The implied meaning acts as a chain-of-thought prompt for understanding indirect responses, assessed using the CircaPlus dataset. Both tasks involve evaluating 2,500 data points.

\noindent{\textbf{Task 4 - Implicature recovery}}
Task 4 differs from tasks 2 and 3 by focusing on implicature recovery in non-polar Question and Answer contexts. In this task,  we present the conversation which is a sequence of QAs ${(Q_1,A_1),(Q_2,A_2),..,(Q_n,A_n)}$  and four choices for the implied meaning of $A_n$. The task for the model is to select an appropriate choice that resolve's the implicature to its explicit form, \textit{i.e}., to perform implicature recovery. We use 2000 data points from the Grice dataset for this task.

While prior tasks have focused on understanding implied meanings in conversations devoid of figurative language, it's important to note that figurative language is a common feature in human communication \cite{lakoff2008metaphors}. Understanding the underlying meanings when such language is used in dialogue is crucial. Therefore, to provide a comprehensive benchmark, we are introducing tasks that focus on understanding implied meanings in conversations where figurative language is present.

\noindent{\textbf{Task 5 and 6 - Agreement detection and Understanding sarcasm}}
Task 5, "Agreement Detection", and Task 6, "Understanding Sarcasm", are both designed to evaluate a language model's ability to comprehend and interpret figurative language within a dialogue. In Task 5, the model is given a conversation between two speakers, a question, and two options: A: Agrees and B: Disagrees. Speaker 1 uses figurative language, and Speaker 2 responds either in agreement or disagreement. The model's objective is to accurately determine if the second speaker concurs with the first. Task 6 flips the roles from Task 5. Here, Speaker 1 makes a statement, and Speaker 2 responds with 'yes', but continues the sentence using figurative language to either agree or disagree (refer to Figure \ref{dataset-diagram} for examples). The model is then tasked with correctly determining if the second speaker is in agreement with the first or is being sarcastic. Modifications are applied to the \cite{figqa} dataset to accommodate both tasks. The evaluation involves 2000 data points for each of the tasks.

\noindent{\textbf{Task 7, 8 and 9 -  Figurative language  understanding using positive and contrastive hints}}
Tasks 7, 8, and 19 are formulated based on the FLUTE dataset \cite{flute}. The FLUTE dataset consists of sentences or premises in figurative language and their corresponding hypotheses in simple language. For each premise, there are two types of hypotheses: one that entails and another that contradicts. Additionally, the dataset includes separate explanations for the entailment and contradiction. In Task 7, the objective is to test if the figurative language is correctly understood. The model must choose between an entailed sentence or a contradictory sentence as the meaning of the premise. In Task 8, the model is provided with an explanation of the entailment, which is referred to as a positive hint as it explains why the entailment option is the correct meaning of the premise. In Task 9, an explanation of the contradictory statement is provided, along with an explanation of why it is not the correct meaning of the figurative sentence. This is considered a contrastive hint. Through these tasks, we aim to test if the models understand the task or if their responses rely on the semantic overlap with the positive hint. The evaluation involves 1770 data points for each of the tasks. 

\noindent{\textbf{Task 10 - Implicature NLI}}
Given that Natural Language Inference (NLI) is a well-established task in the training and evaluation of language models, we have incorporated the NLI task to assess whether the models are capable of making inferences when implicatures are involved. We use 2100 data points from IMPRESS\cite{impres} dataset for this task.

\subsection{Presuppositions}
Presuppositions in a sentence are the underlying assumptions or facts that are implicitly accepted as true by the speaker when making a statement. 

\noindent{\textbf{Task 11 - Presupposition NLI}}
In this task, we approach presupposition verification by framing it as Natural Language Inference (NLI), with an objective akin to that of task 10. We use 1800 data points from IMPRESS \cite{impres} NOPE \cite{nope} dataset for this task. 

\noindent{\textbf{Task 12 - QA over presupposition}}
This task aims to test the ability of the language models on how well they can capture the speaker's assumptions in a dialog. We provide the model with a conversation (set of dialogues between two people), presupposition on the conversation, and two options A. Valid and B. Invalid. The task for the model is to determine if the given presupposition is valid or invalid based on the conversation. We use 2500 data points from the newly annotated DialogAssumptions dataset for this task.

\subsection{Reference}
Deixis, which involves the act of pointing through language, encompasses expressions that are often among the earliest spoken by very young children. These expressions, such as person deixis ('me', 'you'), spatial deixis ('here', 'there'), or temporal deixis ('now', 'then') \citep{yule1996pragmatics}, are indicative of individuals, locations, or times. Deixis is a type of reference closely linked to the speaker's context. 

\noindent{\textbf{Task 13 - Diectic QA}}
This task is designed to access the model's capabilities in resolving references where deictic terms are used. The model is provided with a conversation containing deictic expressions, a polar question regarding reference resolution, and two answer options: A. "Yes" and B. "No.". The model's objective is to accurately determine and provide the correct response to the polar question within the context of the conversation. We selected all the questions and corresponding conversations from the GRICE dataset \citep{grice} that have Yes/No answers. These questions were then filtered using a manually curated list of deictic terms. A total of 2000 data points are used for this task.

\noindent{\textbf{Task 14 - Referential metonymy}}
The task aims to test the model's abilities to understand language use that involves referring to a target object/individual in terms of a distinctive or saliently associated feature. The model is presented with a context featuring metonymic references, along with a question and four possible options. The task requires the model to choose the most suitable option that correctly resolves the reference in response to the question. We use 1100 data points from the newly annotated MetoQA dataset for this task.

\section{Methodology}
\begin{figure}[ht!]
    \centering
    \includegraphics[width=0.6\linewidth]{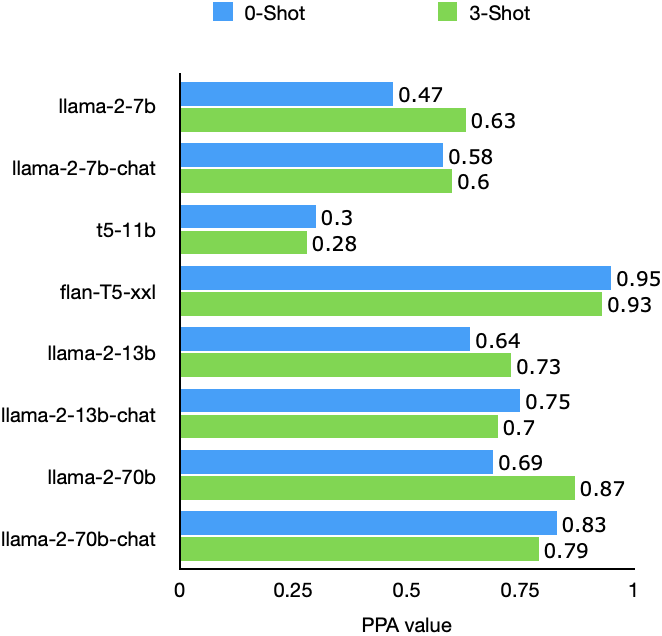}
    \caption{Comparison of various models' multiple choice symbol binding using PPA. Results averaged across Task 4, 11, and 14, representing different pragmatic domains.}
    \label{ppa}
\end{figure}

\begin{figure*}[h!]
    \centering
    \includegraphics[width=0.9\textwidth]{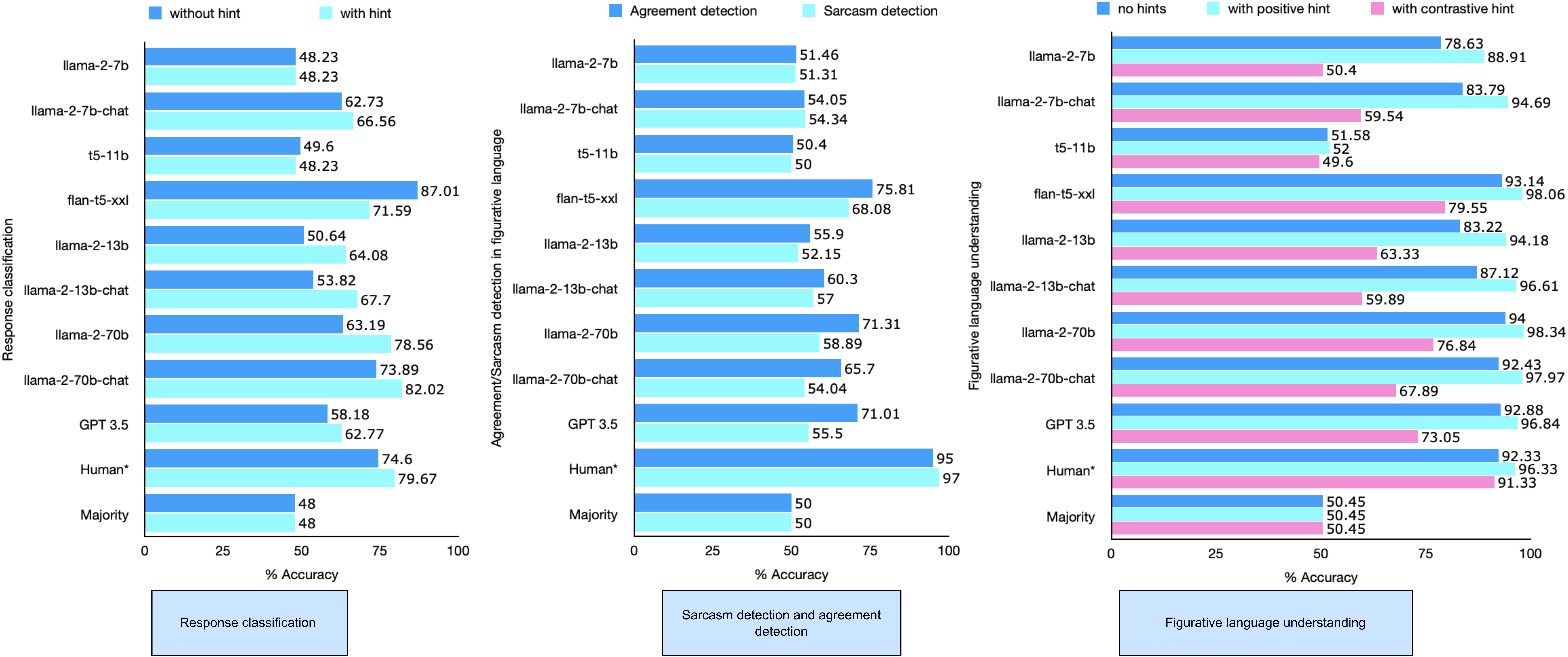}
    \caption{Results (accuracy) for tasks 2 \& 3, tasks 5 \& 6 and tasks 7, 8 \& 9. The results presented in this table are the maximum across all types of evaluations (0-shot and 3-shot Cloze and MCQA) performed on the models.}
    \label{impl_results_2}
\end{figure*}
We have selected two evaluation methods namely length normalized Cloze prompting \cite{DBLP:conf/nips/BrownMRSKDNSSAA20}  and Multiple Choice Prompting (MCP) \cite{RobinsonW23} considering the capabilities of all the models. We have also computed the Proportion of Plurality Agreement (PPA) \cite{RobinsonW23} for all the models to ensure the model's consistency across possible orders of answer options. The results for PPA are presented in Figure \ref{ppa}. We see that vanilla LLMs show improved consistency with a few shots, while instruction-tuned models don't benefit from additional examples. The models under investigation include flan-t5-xxl \citep{flant5}, llama-2 \citep{llama-2}, t5 \cite{t5},  and GPT-3.5 \citet{DBLP:conf/nips/BrownMRSKDNSSAA20}.
\subsection{Sampling for few-shot prompts}
For Zero-shot prompts, all the instances of the data were used as is. For Few-shot prompts, a dev set of 20 examples was created. These 20 examples were selected to ensure a balanced representation of options. For tasks that have unique options for each question, 20 examples were randomly selected from the entire dataset. Depending on the value of \textit{k} for \textit{k}-shot prompt, \textit{k} samples were randomly selected from this dev set. The remaining instances of the data, other than the dev set, were used to evaluate the model.
\subsection{Human evaluation}
To compare the performance of these LLMs with humans, we selected 100 examples from the complete evaluation set for each task. We employed three human evaluators for each task. Each of the 3 human evaluators evaluated these 100 samples for 14 tasks. In total, we have performed 4,200 human evaluations. The samples were chosen to ensure a balanced representation of all option types. The evaluators are fluent English speakers and have graduated from a technical university where English is the medium of instruction. It is important to note that the human evaluation does not reflect expert human reference, but rather random human performance on complex pragmatic tasks. These evaluators are presented with the same prompt as the \textit{0-shot} MCP presented to the LLMs.

\section{Results and Analysis}

\begin{figure*}[h]
    \centering
    \begin{subfigure}{}
        \includegraphics[width=0.74\linewidth]{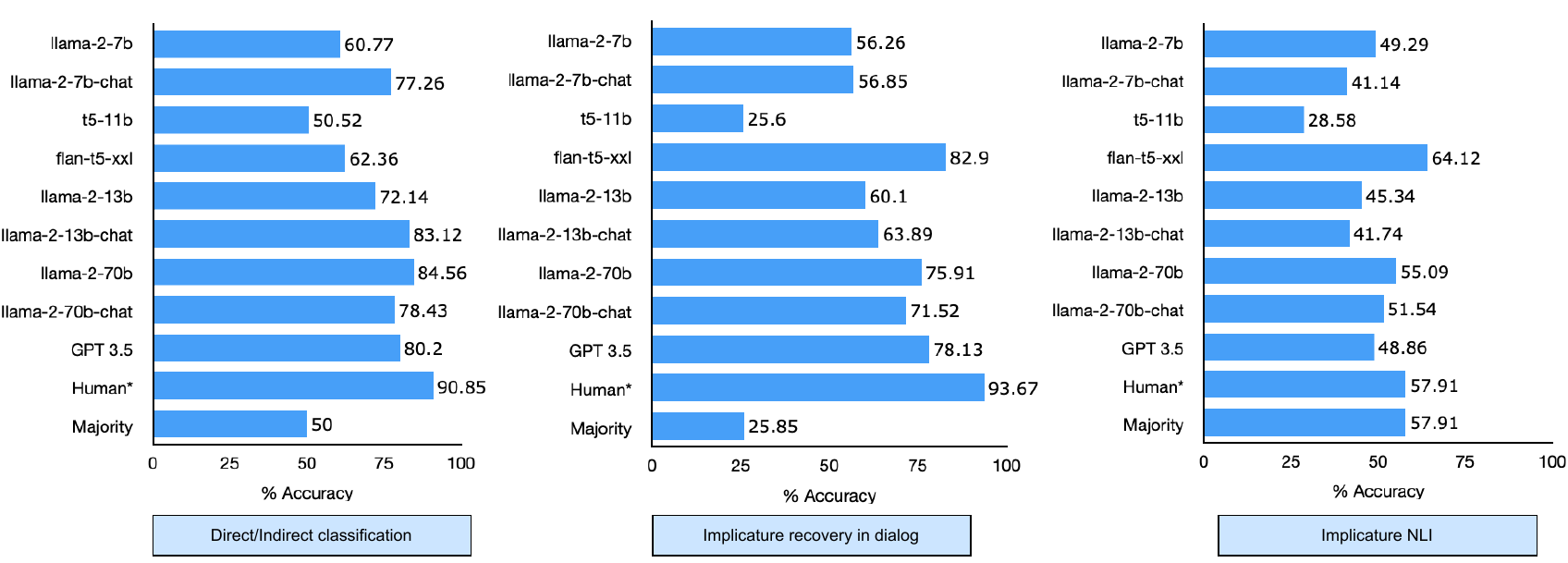}
    \end{subfigure}
    \begin{subfigure}{}
        \includegraphics[width=\linewidth]{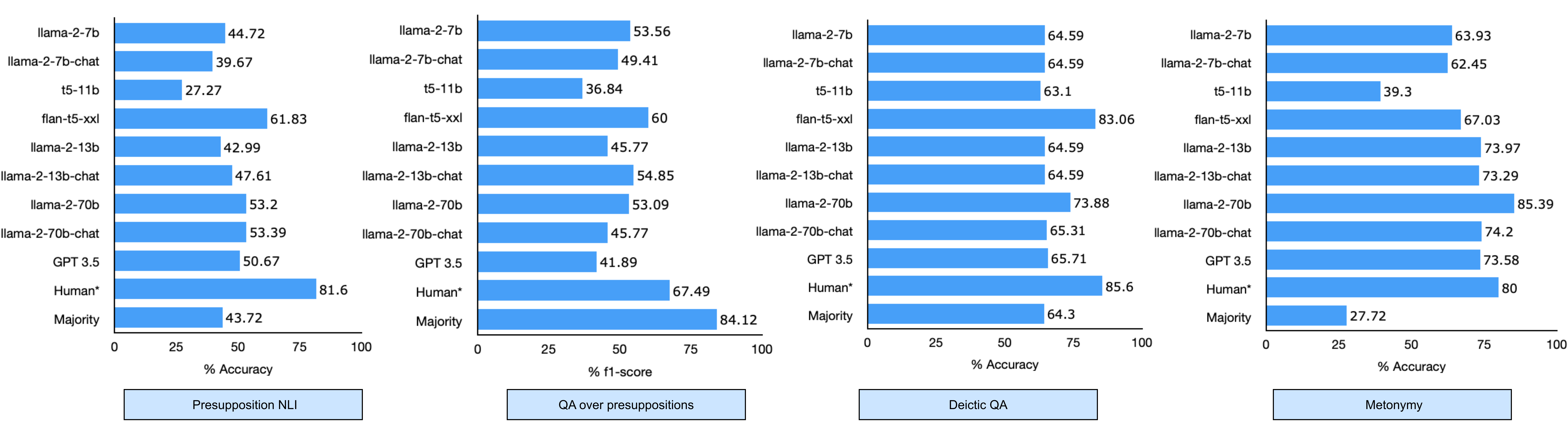}
        \caption{Results for Task 1, 4, 10, 11, 12, 13 and 14. The results presented in this table are the maximum across all types of evaluations (0-shot and 3-shot Cloze and MCQA) performed on the models.}
        \label{pr_results}
    \end{subfigure}
\end{figure*}
We evaluate all the open-source models using both the evaluation methods, i.e. length normalized cloze prompt method and multiple choice prompts.  In each of these methodologies, we do a \textit{zero-shot} evaluation and a \textit{3-shot evaluation}. The OpenAI model is evaluated using MCP. 

\subsection{Results}
The results of our experiments are presented in Figures \ref{impl_results_2}, \ref{pr_results}. Based on these results, we try to address the questions raised in the introduction.

\textbf{How much do LLMs understand what humans mean during conversations?} 
 To evaluate how well LLMs understand what humans intend during conversations, tasks related to implicature and reference offer pertinent insights. We observe that the models perform moderately in the classification of a response as direct or indirect. They struggle to interpret the meaning of the indirect response. Notably, except for the \textit{llama-70b-chat} model, this trend persists across the models evaluated. Furthermore, in this specific task, a slight but noticeable increase in performance is observed across most models when a hint is provided. Interestingly this pattern aligns closely with human performance. The performance trend remains the same in task 4, focusing on resolving implicature in non-polar question-answer scenarios. Even though, NLI is an established task in NLP, it is observed that models perform poorly on making pragmatic inferences. Figure \ref{performance} shows that the average performance on implicature and reference tasks is similar. 
 
 \textbf{Despite operating on the same dataset, do LLMs demonstrate varying task sensitivity?}
While it's known that LLMs are sensitive to the wording of prompts \cite{sensitiveprompt}, this investigation aims to explore their task sensitivity. Specifically, we want to understand how altering the order of speakers asking a different question or giving a different hint impacts the model's performance. Interestingly, LLMs demonstrate stronger performance in agreement detection compared to sarcasm detection (on average there is a $13\%$ performance gap in models $\geq 13b$ parameters) tasks within the same dataset. 
The tasks designed on flute dataset \citep{flute} shed light on the model's susceptibility to distractions. We can observe that with a  change in the hint from positive to contrastive there is a drastic decrease (on an average of $~20\%$) in the accuracy levels for this task across all the models. Interestingly, the inclusion of a positive hint, which has a higher lexical overlap with the correct answer, seems to boost the performance of the model. In contrast, the model's performance appears to decrease when a contrastive hint is introduced. This observed pattern brings into question the pragmatic abilities of these models, suggesting that their understanding and interpretation of language may be more significantly influenced by the presence and nature of linguistic cues than by inherent logic.

\textbf{Does a Model's Scale Correlate with Its Pragmatic Abilities?}
The overall performance depicted in Figure \ref{performance} hints at a possible correlation between a model's scale and its pragmatic capabilities. However, given the model's vulnerability to task sensitivity, even the largest models display perplexity, as previously discussed.  Consequently, concluding that pragmatics is an emergent ability might be premature due to observed inconsistencies, even among models at the extremes of the scale.

\textbf{Do LLMs that are optimized for dialogue use cases exhibit superior pragmatic abilities?}
From the experiments, it is evident that the chat-optimized variants of \textit{llama} slightly outperform the base models on most of the tasks. However, there is a notable performance discrepancy between models like \textit{t5-11B }and \textit{flan-t5-xxl}, with the instruction-tuned \textit{flan-t5-xxl} model approaching near-human-level performance in many of the tasks. This suggests that instruction tuning can significantly enhance a model's ability to handle complex language tasks, bridging the gap toward human-like understanding and processing of language.

\textbf{How do the pragmatic abilities of LLMs compare concerning world knowledge involvement?}
In implicature tasks, excluding task-1 (Direct/Indirect classification) and task-4 (Implicature recovery in dialog context), the other tasks involve a certain degree of world knowledge. While the Metonymy task requires world knowledge, the Deixis task does not. Upon reviewing the outcomes, it becomes evident that the model's below-par performance is not primarily due to a lack of world knowledge. Instead, it appears to stem from a deficiency in their innate pragmatic abilities. This is evident because even in tasks not reliant on world knowledge, like Deixis, the model's performance isn't on par with tasks involving world knowledge. It suggests that the challenge lies more in the model's pragmatic processing rather than their knowledge base.

\textbf{Do they understand the same implied meaning and make the same assumptions as humans?}
The models demonstrate relatively stronger performance in tasks related to implicature and reference, both of which involve inferred meanings from the speaker. However, the models exhibit shortcomings in capturing the speaker's assumptions, known as presuppositions, as evidenced by the results of presupposition tasks (on average there is a performance gap of $\sim 15\%$ between humans and best performing model). Notably, the model's sensitivity to hints and task variations is an important aspect. Human performance remains consistent across sarcasm detection and agreement detection tasks, whereas the models show significant performance discrepancies in these tasks (with an average difference of $13\%$). Similarly, this gap is also observed in tasks concerning figurative language understanding with models showing an average gap of $\sim25\%$ and human performance only differs by $1\%$.

\subsection{Error Analysis}

\begin{figure}[h!]
    \centering
    \includegraphics[width=\linewidth]{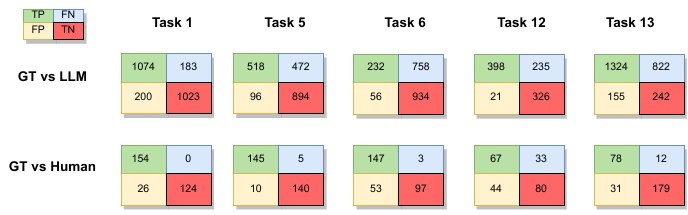}
    \caption{Confusion matrix comparing ground truth with Language Models (LLMs) and ground truth with humans, revealing LLMs' tendency to misclassify positive labels as negatives. Here GT refers to ground truth.}
    \label{fig conf matrix}
\end{figure}

In this section, we look into cases where LLMs fall short in simple pragmatic understanding tasks that humans do with ease. More specifically, we consider the LLaMA-2-70b base model due to its consistently high performance across various tasks and models. For implicature understanding, we see that the model fails to understand the meaning of the response when the response involves complex language phenomena like phrases, expressions, assumptions, or instances where common sense is needed, etc. We compare mistakes of humans and LLMs to see if there is any correlation in pragmatic understanding and if so, is it significant? 
\begin{table}[]
\centering
\resizebox{0.7\columnwidth}{!}{\begin{tabular}{|c|c|c|}
\hline
\rowcolor[HTML]{B0B3B2} 
\textbf{Task No.} & \textbf{GT-Human} & \textbf{Human-LLM} \\ \hline
Task 1            & 0.829             & 0.749 (\textcolor{red}{-0.08})              \\
Task 2            & 0.681             & 0.421 (\textcolor{red}{-0.26})             \\
Task 3            & 0.754             & 0.550 (\textcolor{red}{-0.20})              \\
Task 5            & 0.901             & 0.515 (\textcolor{red}{-0.39})             \\
Task 6            & 0.940              & 0.340 (\textcolor{red}{-0.60})              \\
Task 10           & 0.402             & 0.374 (\textcolor{red}{-0.03})             \\ \hline
Task 11           & 0.565             & 0.269 (\textcolor{red}{-0.30})             \\
Task 12           & 0.350              & 0.327 (\textcolor{red}{-0.02})             \\ \hline
Task 13           & 0.685             & 0.544 (\textcolor{red}{-0.14})             \\ \hline
\end{tabular}}
\caption{Phi coefficient ($\phi$) correlations among Ground Truth (GT), Human evaluator (Human), and LLaMA-2-base-70B (LLM) across 300 examples. Tasks 1-10 examine Implicature, Tasks 11-12 assess Presupposition, and Task 13 focuses on Reference and Deixis. Red text indicates correlation differences between GT-Human and Human-LLM for each task.}
\label{table: MCC comparison}
\end{table}
To see the correlation between human evaluators and LLMs, we report the Phi coefficient $\phi$ (Matthew’s correlation coefficient) in Table \ref{table: MCC comparison} between LLMs (LLaMA-2-70b-base) vs human evaluators (Human-LLM) and compare it with ground truth vs human evaluators (GT-Human). $\phi$ ranges from -1 to 1 where 1 means total agreement, 0 means the predictions are random with respect to the actual values, and -1 means total disagreement. Although we see that for some tasks the correlation values are more than random in Human-LLM, meaning they do make some similar mistakes when compared with GT-Human to see that still there is a large difference and LLMs do not always make the same mistakes as humans. This can be seen in Task 3, where the performance is the same for the LLM and human is the same but there is a correlation gap. This can also be seen in Figure \ref{fig conf matrix} where LLMs do make different mistakes than humans during classification.

For the task of response classification, we see examples where the model thinks that the response is true given some conditions are met but humans do not consider the context as a condition but rather as an auxiliary information. See examples below
\begin{lstlisting}[linewidth=\columnwidth,breaklines=true]
Task 2
Context: X and Y are colleagues leaving work on a Friday at the same time.
X: Have you made dinner plans yet?
Y: I have reservations at the new French place.
Chosen answer: Yes, but with some conditions.

Context: Y has just told X that he/she is thinking of buying a flat in New York.
X: Have you already researched some places?
Y: I plan to discover places by walking around the city.
Chosen answer: Something in the middle
\end{lstlisting}
We also encounter examples where Y’s response is what we call a “polite decline” since there isn't a direct no in the response but an implied No in a tactful manner. 
For understanding implicature in figurative language, we often see examples where metaphors, hyperbole, and tautological statements exist but are in agreement with the speaker.
\begin{lstlisting}[linewidth=\columnwidth,breaklines=true]
Task 6
Speaker_1: The book is a quick, entertaining read
Speaker_2: True, Reading the book is a fun little jog
Chosen answer: Sarcastic disagreement
\end{lstlisting}
We see that in Tasks 5 and 6 the model often confuses agreements with figurative language as sarcastic disagreement but can correctly differentiate sarcastic statements from statements that agree with the speaker, as shown below. Using distractors in figurative language understanding tasks shows how vulnerable LLMs are in their pragmatic abilities. We see that adding a distractor hint in the task confuses LLM and in many cases falls short whereas humans are more robust and see that the hint is contrasting and helps distinguish both meanings of the sentence in the context and choose the correct one. 
\begin{lstlisting}[linewidth=\columnwidth,breaklines=true]
Task 9
Sentence: The ex-slave tasted freedom shortly before she died.
Hint: To taste something means to experience it or enjoy it, while to die before getting something means to never experience it or enjoy it.
Chosen answer: The ex-slave was so close to getting her freedom, but she died before that.
\end{lstlisting}
In instances of presupposition, we observe a recurring pattern where the model erroneously interprets negatives as positives. In the following example, Speaker A expresses frustration about the unsanitary condition of the room, attributing it to the presence of cockroaches. However, the model incorrectly dismisses the notion that being "knee-deep in cockroaches" signifies unhygienic conditions, deeming it an invalid presupposition.
\begin{lstlisting}[linewidth=\columnwidth,breaklines=true]
Task 12
Conversation:
A: I want to change rooms immediately, plus a refund for tonight. 
B:  I'm sorry, sir. Exactly what is the problem? 
A: I'm knee-deep in cockroaches!
Assumption: The room is unhygienic.
Chosen answer: Invalid
\end{lstlisting}
Although LLaMA-2 achieves better results compared to humans in Metonymy understanding, it makes trivial mistakes where humans get it right. But humans fail in cases when reference is one which they are not familiar with, but LLMs due to access to vast and diverse sources of texts get it right.  This task requires common sense and a bit of world knowledge to understand references which humans learn over time. A few examples are given below where the LLM takes the semantic meaning of the reference instead of the pragmatic one.
\begin{lstlisting}[linewidth=\columnwidth,breaklines=true]
Task 14
Context: The chisel sculpted the masterpiece
Question: what does "chisel" refer to?
Chosen answer: Blade

Context: I drive a BMW today 
Question: What does "BMW" stand for?
Chosen answer:  The Brand BMW
\end{lstlisting}
From this error analysis, we find that LLMs don’t make the same mistakes as humans and get confused easily, but more importantly, LLMs fail in trivial cases where humans easily understand the underlying pragmatic answer. More insight into why LLMs fail in such cases is required but we leave that for future research work.

\section{Conclusion}
In this study, we introduce the Pragmatic Understanding Benchmark (PUB) designed to assess pragmatic comprehension in LLMs. We offer a comprehensive analysis, providing insights into various aspects of pragmatic understanding within LLMs. Our findings reveal that pragmatic understanding in LLMs can be enhanced through instruction-tuning of these models. Interestingly, even without specific fine-tuning, language models at scale exhibit equivalent pragmatic understanding. Notably, smaller models, particularly the instruction-tuned variants, outperform their base counterparts, but this advantage diminishes as models scale up, with base and instruction-tuned models showing comparable performance. Despite advancements, LLMs are yet to attain human-level performance, especially in tasks requiring a deep understanding of language context. The observed variability in model performance across different tasks within the same dataset highlights the complexity of achieving human-like pragmatic understanding in LLMs. The PUB benchmark thus provides a clear indication of where LLMs currently stand and the strides still needed to reach human parity in language understanding. We hope that this benchmark will aid researchers in improving LLMs’ conversational abilities with humans.

\section{Limitations}
Our work addresses an important benchmark that can be used to understand and improve the chat capabilities of language models. While we carefully put together a benchmark for evaluation, it's important to note that there might be biases present that may show up in evaluations.
Furthermore, we employed different sampling techniques to avoid evaluation bias for different classes. Although we tried our best to evaluate the models consistently, the models are sensitive to prompt wordings. For the same prompts too, the models are not consistent with the answers when changed the order of options as mentioned in PPA. Therefore there can be slight variations in the performances when trying to reproduce the results. The human evaluation scores reported in the paper are done by graduate students who are proficient in English and language understanding, the results may vary for different sets of human evaluators. The inconsistency of language models is another issue for MCQA results \citep{RobinsonW23}, since inconsistency in answers can lead to false results but until better evaluation methods arrive, we rely on the methods currently used in the paper.

\bibliography{tacl2021}
\bibliographystyle{acl_natbib}

\iftaclpubformat

\onecolumn

\appendix





  

\end{document}